# Objective Evaluation-based High-efficiency Learning Framework for Hyperspectral Image Classification

Xuming Zhang, Jian Yan, Jia Tian, Wei Li, *Senior Member, IEEE*, Xingfa Gu, Qingjiu Tian*

*Abstract*—Deep learning methods have been successfully applied to hyperspectral image (HSI) classification with remarkable performance. Because of limited labelled HSI data, earlier studies primarily adopted a patch-based classification framework, which divides images into overlapping patches for training and testing. However, this approach results in redundant computations and possible information leakage. In this study, we propose an objective evaluation-based high-efficiency learning framework for tiny HSI classification. This framework comprises two main parts: (i) a leakage-free balanced sampling strategy, and (ii) a modified end-to-end fully convolutional network (FCN) architecture that optimizes the trade-off between accuracy and efficiency. The leakage-free balanced sampling strategy generates balanced and non-overlapping training and testing data by partitioning an HSI and the ground truth image into small windows, each of which corresponds to one training or testing sample. The proposed high-efficiency FCN exhibits a pixel-to-pixel architecture with modifications aimed at faster inference speed and improved parameter efficiency. Experiments conducted on four representative datasets demonstrated that the proposed sampling strategy can provide objective performance evaluation and that the proposed network outperformed many state-of-the-art approaches with respect to the speed/accuracy tradeoff. Our source code is available at https://github.com/xmzhang2018.

This study was supported by the National Natural Science Foundation of China (42101321, 61922013), the Project funded by China Postdoctoral Science Foundation (2021M701653), the Major Special Project-the China High-Resolution Earth Observation System (30-Y30F06-9003-20/22), and the National Key R&D Program of China (2019YFE0127300 and 2020YFE0200700). (*Corresponding author: Qingjiu Tian*)
Xuming Zhang, Qingjiu Tian, and Jia Tian are with the International Institute for Earth System Science, Nanjing University, Nanjing 210023, China, and also with the Jiangsu Provincial Key Laboratory of Geographic Information Science and Technology, Nanjing University, Nanjing 210023, China (e-mail: dg21270061@smail.nju.edu.cn, tianqj@nju.edu.cn).
Wei Li is with the School of Information and Electronics, Beijing Institute of Technology, Beijing 100081, China.
Jian Yan and Xingfa Gu are with Aerospace Information Research Institute, Chinese Academy of Sciences, Beijing 100049, China.

## I. Introduction

HYPERSPECTRAL images (HSIs) contain hundreds of narrow bands spanning from the visible to the infrared spectrum, thus forming a 3D hypercube. With this abundant spectral information, each material or object possesses a specific spectral signature similar to a unique fingerprint, serving as its identification. Because of their strong representation ability, HSIs have become economical, rapid, and promising tools for various applications such as medical imaging [1], environmental monitoring [2], and urban development observation [3]. Semantic segmentation (also called pixel-level classification) is one of the most fundamental tasks for these applications.

Over the past several decades, multiple HSI classification methods have been developed. Earlier approaches primarily focused on spectral information mining using machine learning methods, including unsupervised algorithms (e.g. clustering [4]) and supervised algorithms (e.g. support vector machines [5] and random forest [6]). Unsupervised algorithms do not rely on labelled data; however, supervised algorithms are generally preferred because of their superior performance. Nevertheless, the inherent high dimensionality and non-linearity of HSIs limit the performance of supervised algorithms, particularly when labelled samples are limited. Several dimensionality reduction techniques, such as feature selection [7], matrix factorisation [8], and manifold learning [9], have been introduced to project hypercube data into lower-dimensional subspaces by capturing the essential information in the data. Considering the spectral heterogeneity and complex spatial distribution of objects, spatial feature mining has attracted considerable attention [7]. Spatial feature extraction methods, such as grey-level co-occurrence matrix [10], guided filtering [11], and morphological operators [12], have been employed to extract spatial features. Other studies adopted kernel-based methods [13], 3D wavelets [14, 15], and 3D Gabor filters [16] to learn the joint spectral–spatial information for classification. Although these traditional methods have achieved considerable progress, they are limited to shallow features and prior knowledge, thus resulting in poor robustness and generalisation.

Deep learning (DL) can automatically learn high-level representations, overcoming the limitations of traditional feature extraction methods. Over the past several years, DL has enabled computer vision to achieve high performance on many challenging tasks, including object detection [17], scene segmentation [18], and image classification [19]. Various DL techniques have been adopted for HSI classification. A multilayer perceptron was designed as an encoder–decoder structure to extract the deep semantic information of HSIs [20]. Chen *et al.* [21] introduced a deep belief network into HSI classification; three architectures based on this network were designed for spectral, spatial, and spectral–spatial feature extraction. In [22], a stacked autoencoder and convolutional neural network (CNN) were employed to encode spectral and spatial features, respectively, which were then fused for classification. Recurrent neural networks (RNNs) [23] and long short-term memory (LSTM) [24] have been applied to analyse hyperspectral pixels as sequential data. Moreover, because a graph convolutional network can handle graph-structured data by modelling topological relations between samples, it was employed to model the long-range spatial relations of HSIs [25]. Recently, Hong *et al.* [26] proposed a transformer-based architecture to replace CNN- and RNN-based networks. This architecture encodes local spectral representations from multiple adjacent pixels rather than the single band used in the original transformer.

Of these DL algorithms, the CNN generally outperforms the others in HSI classification because of its ability and flexibility to aggregate spectral and spatial–contextual information. The convolutional kernels of a CNN maintain the data's spatial structure during feature extraction. The properties of local connections and shared weights enable a CNN to achieve higher accuracy with fewer parameters.

Many CNN-based methods have been proposed for HSI classification, including patch-based classification and fully convolutional network (FCN)-based classification. Earlier studies [27, 28] primarily focused on the patch-based classification, which assigns the category of a pixel by extracting features from the spatial patch centred at the pixel. However, using this method, redundant computation is inevitable because overlap occurs between adjacent patches, as shown in Fig. 1(a). Many FCN-based approaches [29-31] have been proposed to reduce the computational complexity. They feed the initial HSI cube into the network and perform pixel-to-pixel semantic segmentation. Compared with the patch-based classification, this approach usually produces competitive or superior results with less inference time.

However, unlike datasets in computer vision that contain thousands of labelled images, HSI datasets often include only one partially labelled image. Almost all the aforementioned methods employ the random sampling strategy, where the training and testing samples are randomly selected from the same image, thus resulting in the feature extraction space of training and testing data overlaps, as shown in Fig. 1(b). Consequently, in the training stage, information from the testing data is used to train the network, leading to exaggerated results [32]. Similarly, FCN-based approaches may lead to higher training–testing information leakage; thus, their performance and generalisability results are questionable because they violate the fundamental assumption of supervised learning [32]. Although several new sampling strategies [32, 33] have been proposed to avoid training–testing information leakage, other limitations, e.g. imbalanced sampling results in certain categories for which all data is selected as the testing or training set, may emerge.

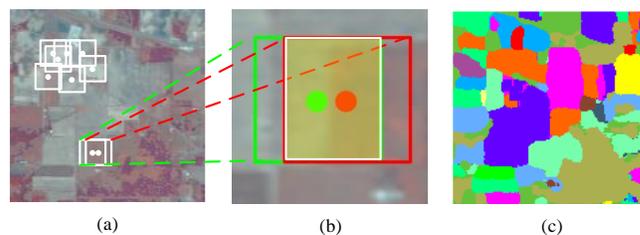

**Fig. 1.** Demonstration of the traditional sampling strategy, which results in (a) overlap between adjacent patches, (b) overlap between the training and testing data, and (c) blurred boundaries of the classification map. In (a), dots represent the central pixels of the corresponding patches with white borders. In (b), green and red dots represent the training and testing pixels of the corresponding patches, respectively.

To address the abovementioned limitations, we propose an objective evaluation-based high-efficiency learning (OEEL) framework for HSI classification and objective performance evaluation. First, to ensure balanced sampling and no training-test information leakage, a leakage-free balanced sampling strategy is proposed to generate training and testing samples. Then, a high-efficiency FCN is designed to learn discriminative spectral–spatial features from the generated samples for effective and efficient data classification.

The primary contributions of this study are summarised as follows:

1) To achieve fast classification and objective evaluation, the OEEL framework is proposed for tiny HSI dataset classification. This framework includes a sampling strategy and an FCN-based architecture.
2) Since the existing sampling strategies are not suitable for limited labelled HSI datasets, four principles that should be considered for effective sampling strategy design are proposed. As per these principles, we design a leakage-free balanced sampling strategy for tiny datasets to generate balanced training and testing data without information leakage, ensuring objective and accurate evaluation.
3) A high-efficiency FCN architecture with high inference and parameter efficiency is developed to extract spectral–spatial features in the small-sample cases. Experiments on four publicly available datasets demonstrate that the proposed architecture is faster and more accurate than previous methods.

The remainder of this study is organised as follows. In Section II, we review patch-based classification, FCN-based classification, and the sampling strategy for HSIs. In Section III, we describe the proposed framework; and in Section IV, we present the experiments and results. In Section V, we present the network analysis, while in Section VI, we provide concluding remarks.

## II. RELATED WORK

### A. *Patch-based Classification*

Most previous studies [27, 28] employed a patch-based classification framework to perform spatial feature extraction and classifier training. In this framework, patches centred at the sample pixels are generated from the original image. These patches are then fed to networks for feature extraction or classification. An end-to-end network was proposed that takes 3D patches as input and outputs a specific label for each patch in its last fully connected (FC) layer [28]. Another end-to-end two-dimensional CNN [34] was proposed that employs $1 \times 1$ convolution kernels to mine spectral information and uses global average pooling to replace FC layers to prevent overfitting. Santara *et al*. [30] proposed a band-adaptive spectral–spatial feature learning neural network to address the curse of dimensionality and spatial variability of spectral signatures. This neural network divides 3D patches into sub-cubes along the channel dimension to extract band-specific spectral–spatial features. To enhance the learning efficiency and prevent overfitting, a deeper and wider network with residual learning was proposed, called contextual deep CNN [35], in which a multi-scale filter bank is employed to jointly exploit spectral–spatial information.

Two-branch CNN-based architectures have been proposed to better aggregate spectral–spatial information [22, 24, 32]. They employ two-dimensional CNNs and other algorithms (e.g. 1D CNN, stacked autoencoder, and LSTM) to encode spatial and spectral information, respectively, and then fuse the outputs for classification. Another type of spectral–spatial-based CNN architecture employs 3D CNN to extract joint spectral–spatial features for HSI classification [28, 36]. For instance, the spectral–spatial residual network (SSRN) [37] uses spectral and spatial residual blocks consecutively to learn spectral and spatial information from raw 3D patches. A fast, dense spectral–spatial convolution framework [38] was proposed that uses residual blocks with $1 \times 1 \times d\,(d>1)$ and $a \times a \times 1\,(a>1)$ convolution kernels to sequentially learn spectral and spatial information.

Recently, networks have introduced attention mechanisms to adaptively emphasise informative features [27]. The squeeze-and-excitation (SE) module [39], which uses global pooling and FC layers to generate channel attention vectors, was adopted in [40, 41] to recalibrate the spectral feature responses. The convolutional block attention module [42] was adopted in [43], where the spatial branch appended a spatial-wise attention module while the spectral branch appended a channel-wise attention module to extract the spectral and spatial features in parallel. Similarly, the position self-attention module and the channel self-attention module proposed in [18] were introduced into a double-branch dual-attention mechanism network (DBDA) [44] to refine the extracted features of HSIs. In [27], a spatial self-attention module was designed for the patch-based CNN to enhance spatial feature representation related to the centre pixel.

Although the above patch-based classification methods achieved high performance, it is unclear whether this was attributed to the improved performance of the methods or the training–testing information leakage [32]. Furthermore, redundant computation of overlapping regions of adjacent patches is inevitable using these methods.

*B. FCN-based Classification*

Many FCN-based frameworks have been developed to mitigate redundant computation caused by overlap between adjacent patches. The spectral–spatial fully convolutional network (SSFCN) [30] takes the original HSI cube as input and performs classification in an end-to-end, pixel-to-pixel manner. A depthwise separable fully convolutional residual network [31] was proposed to efficiently extract features. A deep FCN with an efficient nonlocal module [45] was proposed that takes the entire HSI as input and uses an efficient non-local module to capture long-range contextual information. To exploit global spatial information, Zheng *et al*. [29] proposed a fast patch-free global learning framework that includes a global stochastic stratified sampling strategy and an encoder–decoder-based FCN (FreeNet). However, this framework does not perform well with imbalanced sample data. A spectral–spatial dependent global learning (SSDGL) framework [46] was developed for classification because of imbalanced and insufficient HSI data.

Although these FCN-based classification frameworks alleviate redundant computation and achieve significant performance gains, they may lead to higher training–testing information leakage. This is because they use the same image for both training and testing, thus leading to overlap and interaction between the feature extraction space of the training and testing data.

*C. Sampling Strategy*

The aforementioned training–testing information leakage not only leads to a biased evaluation of spatial classification methods but may also distort the boundaries of objects, as shown in Fig. 1(c). Therefore, the pixel-based random sampling strategy inadvertently affects feature learning and performance evaluation.

To address these limitations, several new sampling strategies have been proposed. A controlled random sampling strategy was designed to reduce the overlap between training and testing samples [32]. In particular, this strategy randomly selects a labelled pixel from each unconnected partition as the seed and then extends the region from the seed pixel to generate training data. Finally, pixels in the grown regions are selected as training data, and the remaining pixels are selected as testing data. This sampling strategy dramatically reduces the overlap between training and testing data; however, it does not eliminate it because pixels at the boundaries of each training area still overlap with the testing data.

Nalepa *et al*. [47] proposed to divide HSIs into fixed-size patches without overlapping and then randomly select some patches as the training set. The method proposed in [33] only selects training samples from multi-class blocks following a specific order. Nevertheless, both of these methods may suffer from severe sample imbalance; i.e., there may be certain categories for which all data is selected as the testing or training set. The former causes the trained model to fail to recognise these categories, while the latter results in a lack of testing samples for the evaluation of these categories. Furthermore, these methods disregard boundary pixels, where a patch cannot be defined. Therefore, the significant loss of samples together with the scarcity of samples may cause overfitting.

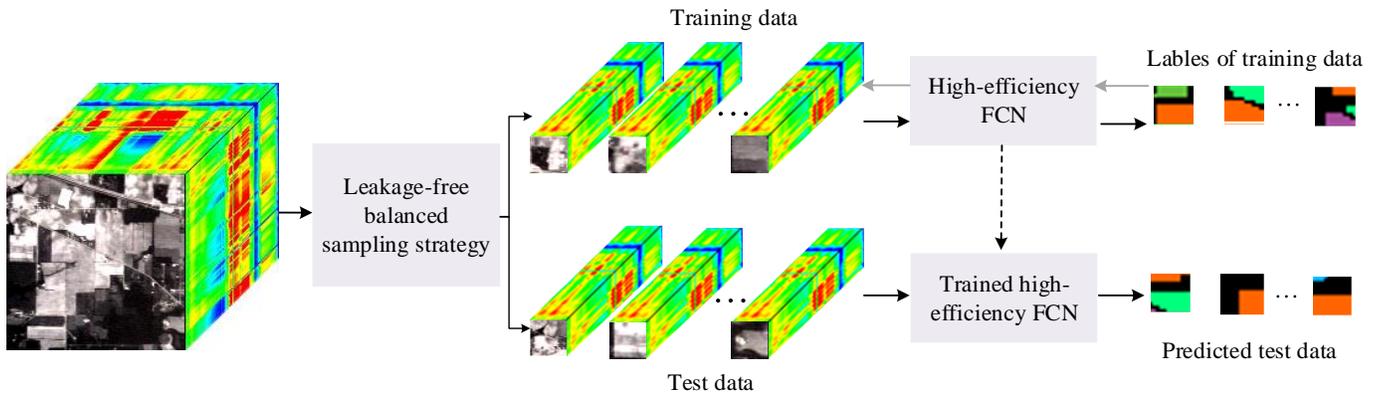

**Fig. 2.** Overview of the proposed OEEL framework. The framework includes two core components: a leakage-free balanced sampling strategy and a high-efficiency FCN.

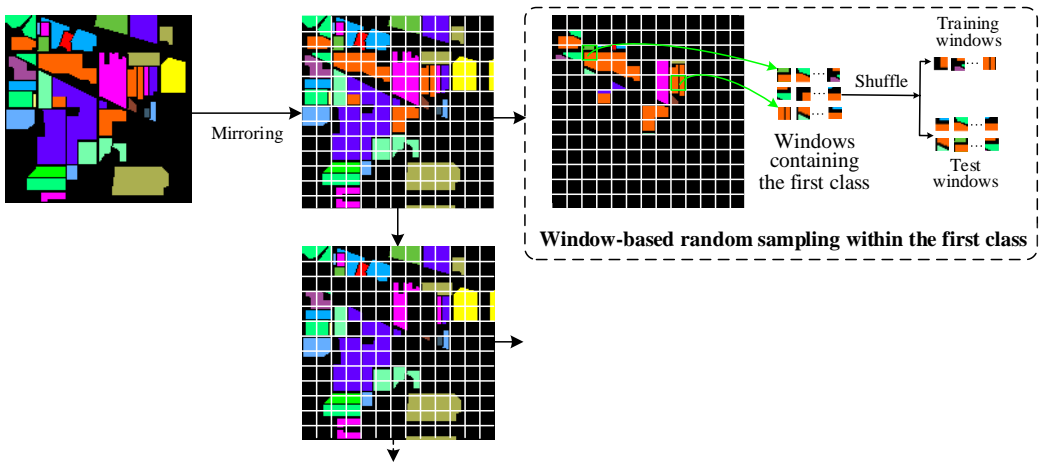

**Fig. 3.** Flowchart of the proposed leakage-free balanced sampling strategy. The operation process of an HSI is the same as that of the ground truth. For convenience, only the ground truth operation process is presented.

### III. METHOD

In this section, the OEEL framework is presented. As shown in Fig. 2, it comprises two main steps. First, the leakage-free balanced sampling strategy divides the HSI cube into non-overlapping training and testing data. Second, the generated training and testing data are used to train and test the proposed high-efficiency FCN for data classification, respectively. The relevant details of both steps are described below.

#### A. Leakage-free Balanced Sampling Strategy

As discussed in Section II.C, Because of limited labelled HSI data, the commonly used sampling strategy exaggerates the classification results because of training–test information leakage. Although several new sampling strategies have been proposed to solve this problem, other limitations may emerge. Therefore, we derived several basic principles for effective sampling strategy design based on these observations and empirical studies [32, 33]: P1) balanced sampling to ensure that all categories are present in both training and testing sets; P2) samples should be maximally utilised; P3) regions that contribute to feature extraction from the training data cannot be used for testing to satisfy the assumption of independence; and P4) random sampling is used to avoid biased estimates.

As per these principles, we designed a leakage-free balanced sampling strategy, as shown in Fig. 3. Because many spatial-based methods require square patches as input, the HSI and its corresponding ground truth must be divided into square windows of equal size. To satisfy P1, the window size should ensure each class in at least two windows, and there is a trade-off between the window size and the number of windows. To satisfy P2, we first mirror the pixels of the right and bottom boundaries outward to create the corresponding windows, as shown in the first step of Fig. 3. Once the border pixels are mirrored, the HSI and its ground truth are split in disjoint windows.

To satisfy P1 and P3-4, we perform window-based random sampling within each category in order. As shown in the dotted box of Fig. 3, windows containing the first class are collected; then, a predefined proportion of windows are randomly selected for training while the remaining windows are used for testing. Then, the corresponding positions of windows containing the first class are set to zero in the HSI and its ground truth, which are then used to collect the next category (P3). This process is repeated until sampling is complete for all categories.

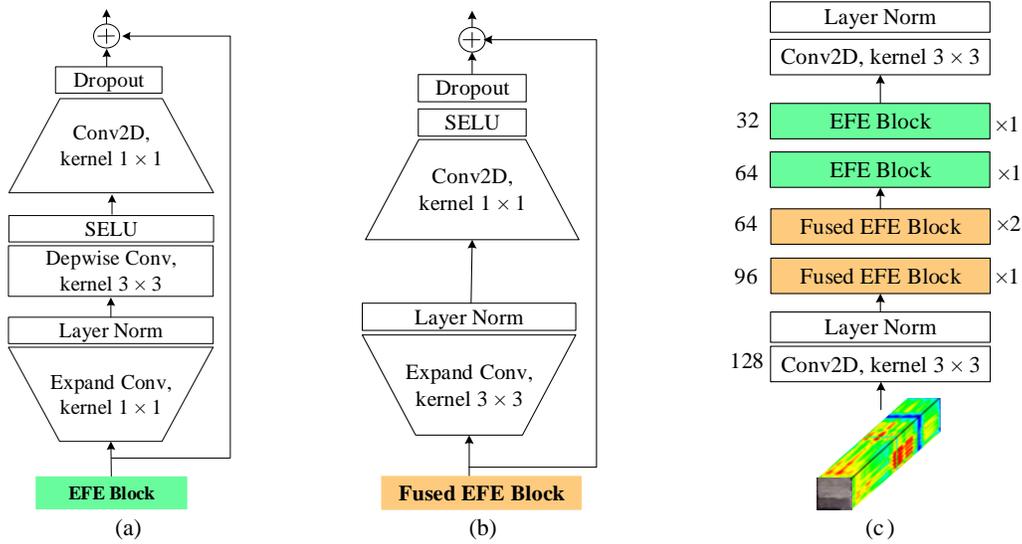

**Fig. 4.** High-efficiency FCN architecture designed for tiny HSI classification. (a) EFE block. (b) Fused EFE block. (c) High-efficiency FCN embedded with EFE blocks and fused EFE blocks, where the number of output channels and repetitions per block is listed on the left and right sides, respectively.

Because of the limited training windows, it is necessary to perform data augmentation to avoid overfitting. As performed in most previous studies [27, 48], each training window is randomly rotated between 0° and 360° and horizontally or vertically flipped. We add noise or change the brightness of training windows to enhance the robustness of approaches under various conditions such as different sensors, light changes, and atmospheric interference.

A summary of the proposed sampling strategy is provided in Algorithm 1. It follows all of the abovementioned principles, enabling the accurate and objective performance evaluation of approaches. Note that our sampling strategy applies not only to HSI but also to other data, particularly data with imbalanced categories.

### B. High-Efficiency FCN

Recent methods, such as the fast, dense spectral–spatial convolution framework [38] and DBDA [44], aim to improve the accuracy and reduce the training time of networks. SSFCN [33], FreeNet [29], and SSDGL [46] are trained and tested on an entire image to find a good trade-off between efficiency and performance. However, their improvement comes at the cost of overestimating results and blurring object boundaries because of information leakage.

To overcome the information leakage, this subsection proposes a high-efficiency FCN that is also optimised for faster inference speed and higher parameter efficiency. It includes two main blocks—the efficient feature extraction (EFE) block and the fused efficient feature extraction (fused EFE) block—which are described as follows.

*1) EFE block:* Because the depthwise convolution [49] has fewer parameters and floating-point operations (FLOPs) compared to regular convolutions, it has been introduced to MBConv [19] to achieve higher parameter efficiency. MBConv is defined by a $1 \times 1$ expansion convolution followed by $3 \times 3$ depthwise convolutions, an SE module, and a $1 \times 1$ projection layer. Its input and output are connected by a residual connection when they have the same number of channels. MBConv attaches batch normalisation (BN) and a sigmoid linear unit (SiLU) activation function to each convolutional layer.

To increase the network efficiency, we first replace SiLU with the scaled exponential linear unit (SELU). SELU exhibits self-normalising properties, which are faster than external normalisation, confirming that the network converges faster. The SELU activation function is defined as follows:

$$selu(x) = \lambda \begin{cases} x & if\ x > 0, \\ \alpha e^x - \alpha & if\ x \leq 0, \end{cases} \quad (1)$$

---

**Algorithm 1** Leakage-free Balanced Sampling Strategy

**Require:** Hyperspectral image $H$, corresponding ground truth $T$, window size, and sampling rate per class
**Preprocessing:** Mirror the pixels of the right and the lower borders outward
**Segmentation:** Split the $H$ and the $T$ into disjoint windows
**for** each class $c$ in $H$ **do**
    Collect the windows containing the $c$-th class from $H$ and $T$
    Count the number of windows
    Randomly select a predefined sampling rate of windows as the training set and the rest as the test set
    Set the corresponding position of these widows as zero at the $H$ and $T$
**end for**
These training subsets and test subsets are combined to form the training set and test set, respectively. Apply data augmentation on the training set.

where $x$ is the input, $\alpha$ and $\lambda(\lambda>1)$ are hyperparameters, and $e$ denotes the exponent. SELU reduces the variance for negative inputs and increases that for positive inputs, thereby preventing vanishing and exploding gradients. Moreover, it emits outputs of zero mean and unit variance. Therefore, SELU converges faster and more accurately than SiLU, thus leading to better generalisation [50].

Layer normalisation (LN) has been used in ConvNeXt [51], achieving slightly better performance than BN for different application scenarios. Following the same optimisation strategy as in [51], we substitute BN with LN in our network.

LN and activation function operations cost considerable time [52]. ConvNeXt uses fewer LN and activation functions, achieving better results. Therefore, we use fewer LN and SELU activation functions to improve accuracy and efficiency. As shown in Fig. 4(a), the LN and activation function are only attached after the expansion convolution and depthwise convolution, respectively. Furthermore, because of the high computational cost of FC layers in SE, the SE module is removed. The results in Section V-B demonstrate that this modification not only improves the training speed and parameter efficiency but also improves the classification performance.

Fig. 4(a) shows the detailed architecture of the EFE block. It comprises an expansion convolution with LN, followed by 3 × 3 depthwise convolutions with the SELU activation function and a 1 × 1 projection layer. The expansion ratio of the first 1 × 1 convolution is then set to 2. Similarly, the input and output of the EFE block are connected via a residual connection when they have the same number of channels.

*2) Fused EFE block:* Because depthwise convolutions cannot fully utilise modern accelerators, Fused-MBConv has been proposed [19]. It replaces the 3 × 3 depthwise convolutions and 1 × 1 expansion convolution in MBConv with a single regular 3 × 3 convolution. To improve the training speed, we follow the method of Fused-MBConv to replace the 1 × 1 expansion convolution and 3 × 3 depthwise convolutions in the EFE block with a single regular 3 × 3 convolution, as shown in Fig. 4(b). Similarly, LN and SELU are only appended after the 3 × 3 convolution and 1 × 1 convolution, respectively. Similar to the EFE block, the expansion ratio of the first 1 × 1 convolution is set to 2.

*3) High-Efficiency FCN:* It has been demonstrated that depthwise convolutions are slow at the early stages but effective in deep layers [19]. Thus, the EFE block is placed in the shallow layers. After incorporating the EFE and fused EFE blocks in the network, the high-efficiency FCN architecture can be developed, as shown in Fig. 4(c), where the number of repetitions and output channels is presented to the left and right of each block, respectively. The network aims to learn a mapping of $\mathbf{X}_i \in \mathbb{R}^{h \times w \times B} \to \mathbf{Y}_i \in \mathbb{R}^{h \times w \times K}$ for classification, where $h \times w$ and $B$ are the spatial size and the number of bands of $\mathbf{X}$, respectively, and $K$ is the number of categories to be classified.

In our network, the number of channels starts at the maximum value and decreases as the layer deepens. We refer to this operation as *inverted channels*. HSIs with abundant spectral information inevitably contain a high degree of redundancy between bands. *Inverted channels* can allow the network to learn additional valuable information from the redundant bands.

There are no pooling layers in the entire network. The reasons for this are primarily two-fold. First, the pooling operations are performed on summarised rather than positioned features, thus making the network more invariant to spatial transformations. In turn, the spatial invariance limits the accuracy of semantic segmentation. Second, the pooling operation is primarily used to reduce the amount of computation by reducing the spatial dimensions of the feature maps. This operation results in significant spatial information loss and may blur the boundaries of land covers, particularly when the input window size is small. Moreover, our task is pixel-wise classification; thus, the network output should have the same spatial dimension as the input. Consequently, we do not adopt any downsampling operations.

After the high-efficiency FCN is constructed, its parameters are initialised and trained end to end. The performance of the proposed FCN is presented and discussed in Section IV.

## IV. EXPERIMENTS

This section describes the experimental datasets and settings, including the comparison methods, evaluation metrics, and parameter settings. Moreover, quantitative and qualitative analyses of experimental results are presented.

### A. Description of Datasets

We performed experiments on the following four datasets: *Indian Pines* (IP), *Pavia University* (PU), *Salinas* (SA), and *University of Houston* (UH).

The IP dataset was collected in 1992 by the Airborne Visible/Infrared Imaging Spectrometer (AVIRIS) sensor over northwestern Indiana, USA, which is an agricultural area with irregular forest regions and crops of regular geometry. The dataset has 145 × 145 pixels with a spatial resolution of 20 m. Each pixel has 224 spectral bands ranging from 0.4 to 2.5 μm. After discarding 24 noisy and water absorption bands (i.e. 104–108, 150–163, and 220), 200 bands were used for classification. The ground truth has 16 land-cover classes. Fig. 5(a) summarises the class name and number of samples. The spatial distribution of the training data is provided in Fig. 5(b), which was produced using the proposed sampling strategy.

The PU dataset covering the University of Pavia, Northern Italy, was collected by the Reflective Optics System Imaging Spectrometer sensor in 2001. The dataset is a 610 × 340 × 115 data cube with a spatial resolution of 1.3 m and a wavelength range of 0.43–0.86 μm. Before the experiments, the number of spectral bands was reduced to 103 because of the removal of water absorption bands. The scene is an urban environment characterised by natural objects and shadows, where nine land-cover classes are labelled. Detailed information about this dataset is provided in Fig. 6.

| ID | Colour | Land Cover Type | Train | Test | Total |
|---|---|---|---|---|---|
| C1 | | Alfalfa | 23 | 23 | 46 |
| C2 | | Corn-notill | 312 | 1116 | 1428 |
| C3 | | Corn-mintill | 241 | 589 | 830 |
| C4 | | Corn | 61 | 176 | 237 |
| C5 | | Grass-pasture | 104 | 379 | 483 |
| C6 | | Grass-trees | 191 | 539 | 730 |
| C7 | | Grass-pasture-mowed | 16 | 12 | 28 |
| C8 | | Hay-windrowed | 128 | 350 | 478 |
| C9 | | Oats | 8 | 12 | 20 |
| C10 | | Soybean-notill | 215 | 757 | 972 |
| C11 | | Soybean-mintill | 577 | 1878 | 2455 |
| C12 | | Soybean-clean | 159 | 434 | 593 |
| C13 | | Wheat | 45 | 160 | 205 |
| C14 | | Woods | 308 | 957 | 1265 |
| C15 | | Buildings | 104 | 282 | 386 |
| C16 | | Stone-Steel-Towers | 43 | 50 | 93 |
| | | Total | 2535 | 7714 | 10249 |

(a)

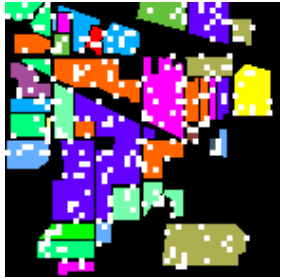

(b)

**Fig. 5.** IP dataset. (a) Land cover type and sample settings. (b) Spatial distribution of training samples (white windows).

| ID | Colour | Land Cover Type | Train | Test | Total |
|---|---|---|---|---|---|
| C1 | | Asphalt | 361 | 6270 | 6631 |
| C2 | | Meadows | 801 | 17848 | 18649 |
| C3 | | Gravel | 100 | 1999 | 2099 |
| C4 | | Trees | 142 | 2922 | 3064 |
| C5 | | Painted metal sheets | 115 | 1230 | 1345 |
| C6 | | Bare Soil | 309 | 4720 | 5029 |
| C7 | | Bitumen | 57 | 1273 | 1330 |
| C8 | | Self-Blocking Bricks | 248 | 3434 | 3682 |
| C9 | | Shadows | 60 | 887 | 947 |
| | | Total | 2193 | 40583 | 42776 |

(a)

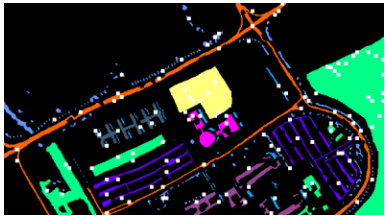

(b)

**Fig. 6.** PU dataset. (a) Land cover type and sample settings. (b) Spatial distribution of training samples (white windows).

| ID | Colour | Land Cover Type | Train | Test | Total |
|---|---|---|---|---|---|
| C1 | | Brocoli_green_weeds_ | 172 | 1837 | 2009 |
| C2 | | Brocoli_green_weeds_ | 26 | 3700 | 3726 |
| C3 | | Fallow | 135 | 1841 | 1976 |
| C4 | | Fallow_rough_plow | 79 | 1315 | 1394 |
| C5 | | Fallow_smooth | 107 | 2571 | 2678 |
| C6 | | Stubble | 150 | 3809 | 3959 |
| C7 | | Celery | 70 | 3509 | 3579 |
| C8 | | Grapes_untrained | 307 | 1096 | 11271 |
| C9 | | Soil_vinyard_develop | 163 | 6040 | 6203 |
| C10 | | Corn_senesced_green | 149 | 3129 | 3278 |
| C11 | | Lettuce_romaine_4wk | 70 | 998 | 1068 |
| C12 | | Lettuce_romaine_5wk | 81 | 1846 | 1927 |
| C13 | | Lettuce_romaine_6wk | 99 | 817 | 916 |
| C14 | | Lettuce_romaine_7wk | 114 | 956 | 1070 |
| C15 | | Vinyard_untrained | 237 | 7031 | 7268 |
| C16 | | Vinyard_vertical_trelli | 162 | 1645 | 1807 |
| | | Total | 2121 | 5200 | 54129 |

(a)

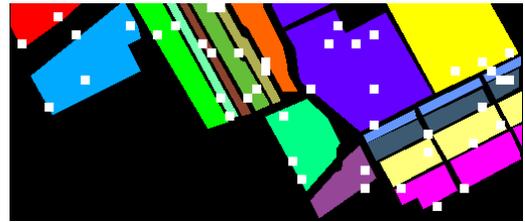

(b)

**Fig. 7.** SA dataset. (a) Land cover type and sample settings. (b) Spatial distribution of training samples (white windows).

| ID | Colour | Land Cover Type | Train | Test | Total |
|---|---|---|---|---|---|
| C1 | | Healthy grass | 250 | 1001 | 1251 |
| C2 | | Stressed grass | 201 | 1053 | 1254 |
| C3 | | Artificial turf | 244 | 453 | 697 |
| C4 | | Evergreen trees | 226 | 1018 | 1244 |
| C5 | | Deciduous trees | 284 | 958 | 1242 |
| C6 | | Bare earth | 57 | 268 | 325 |
| C7 | | Water | 262 | 1006 | 1268 |
| C8 | | Residential buildings | 222 | 1022 | 1244 |
| C9 | | Non-residential | 206 | 1046 | 1252 |
| C10 | | Roads | 217 | 1010 | 1227 |
| C11 | | Sidewalks | 180 | 1055 | 1235 |
| C12 | | Crosswalks | 156 | 1077 | 1233 |
| C13 | | Major thoroughfares | 108 | 361 | 469 |
| C14 | | Highways | 21 | 407 | 428 |
| C15 | | Railways | 171 | 489 | 660 |
| | | Total | 2805 | 12224 | 15029 |

(a)

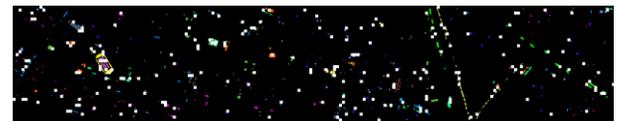

(b)

**Fig. 8.** UH dataset. (a) Land cover type and sample settings. (b) Spatial distribution of training samples (white windows).

The SA dataset was recorded by the AVIRIS sensor over several agricultural fields in Salinas Valley, California, USA. It contains 512 × 217 pixels with a spatial resolution of 3.7 m per pixel. Each pixel has 224 spectral bands in the spectral range of 0.36–2.5 μm. As in the case of the IP dataset, 20 noise and water absorption bands were discarded before the experiments. As summarised in Fig. 7(a), 16 land-cover classes were defined. Fig. 7(b) shows the spatial distribution of the training data.

The UH dataset covers an urban area including the University of Houston campus and neighbouring areas. It was collected by the National Centre for Airborne Laser Mapping in June 2012. It has 144 spectral bands in the wavelength range of 0.38–1.05 μm. Furthermore, the spatial dimension and resolution of this scene are 349 × 1905 and 2.5 m, respectively. There are 15 classes in this scene, and detailed information about this dataset is presented in Fig. 8.

TABLE I
COMPARISON OF CLASSIFICATION ACCURACY OF DIFFERENT METHODS ON THE IP DATASET

| Metrics | SSRN | DBDA | SS3FCN | FreeNet | SSDGL | EfficientNetV2 | ConvNeXt | Proposed |
|---|---|---|---|---|---|---|---|---|
| OA (%) | 77.68 ± 0.85 | 78.37 ± 1.00 | 76.67 ± 1.43 | 81.76 ± 1.25 | 82.07 ± 0.56 | 79.18 ± 0.72 | 81.08 ± 0.34 | **84.72 ± 0.59** |
| AA (%) | 73.49 ± 0.66 | 73.68 ± 0.62 | 76.92 ± 1.12 | 83.72 ± 1.09 | 83.26 ± 0.72 | 80.04 ± 1.09 | 79.89 ± 0.57 | **84.92 ± 0.84** |
| $\kappa \times 100$ | 74.42 ± 0.99 | 75.26 ± 1.07 | 73.29 ± 1.32 | 79.23 ± 1.17 | 79.53 ± 0.61 | 76.16 ± 0.83 | 78.41 ± 0.39 | **82.53 ± 0.63** |
| C1 | 97.72 | 40.84 | 72.31 | 99.16 | 98.89 | 96.70 | 94.61 | **99.35** |
| C2 | 77.83 | 75.5 | 74.18 | 82.45 | 84.33 | 79.40 | 85.13 | **89.11** |
| C3 | 60.87 | 60.92 | 63.45 | **71.30** | 70.93 | 59.23 | 65.95 | 64.24 |
| C4 | 36.5 | 37.21 | 43.27 | **62.83** | 44.50 | 42.26 | 52.36 | 46.34 |
| C5 | 78.56 | 75.13 | 82.45 | 94.59 | 80.47 | 84.65 | 91.06 | **95.03** |
| C6 | 96.3 | **97.14** | 89.85 | 91.77 | 92.34 | 92.43 | 93.73 | 95.81 |
| C7 | 79.64 | 95.07 | 94.78 | **100** | **100** | **100** | **100** | **100** |
| C8 | 96.38 | 97.97 | 91.85 | 82.32 | 95.64 | 92.90 | 95.82 | **97.76** |
| C9 | 11.64 | 28.4 | **87.12** | 86.67 | 86.67 | 73.81 | 38 | 66.67 |
| C10 | 75.78 | 67.77 | 68.95 | 86.61 | 83.10 | 81.14 | 86.80 | **88.60** |
| C11 | 75.36 | 77.31 | 75.03 | 77.88 | 79.03 | 77.83 | 74.82 | **79.95** |
| C12 | 69.99 | 78.09 | 67.69 | 74.66 | 72.19 | 61.65 | 58.00 | **81.77** |
| C13 | 93.34 | 93.57 | 94.79 | **98.05** | 97.21 | 96.98 | 94.93 | 95.27 |
| C14 | 95.05 | 94.22 | 96.72 | 94.36 | 97.27 | 95.76 | 96.83 | **97.49** |
| C15 | 39.95 | 63.31 | 50.54 | 38.62 | 50.46 | 49.63 | 53.81 | **64.62** |
| C16 | 90.92 | 96.95 | 77.80 | 98.29 | **99.14** | 96.22 | 96.41 | 96.58 |

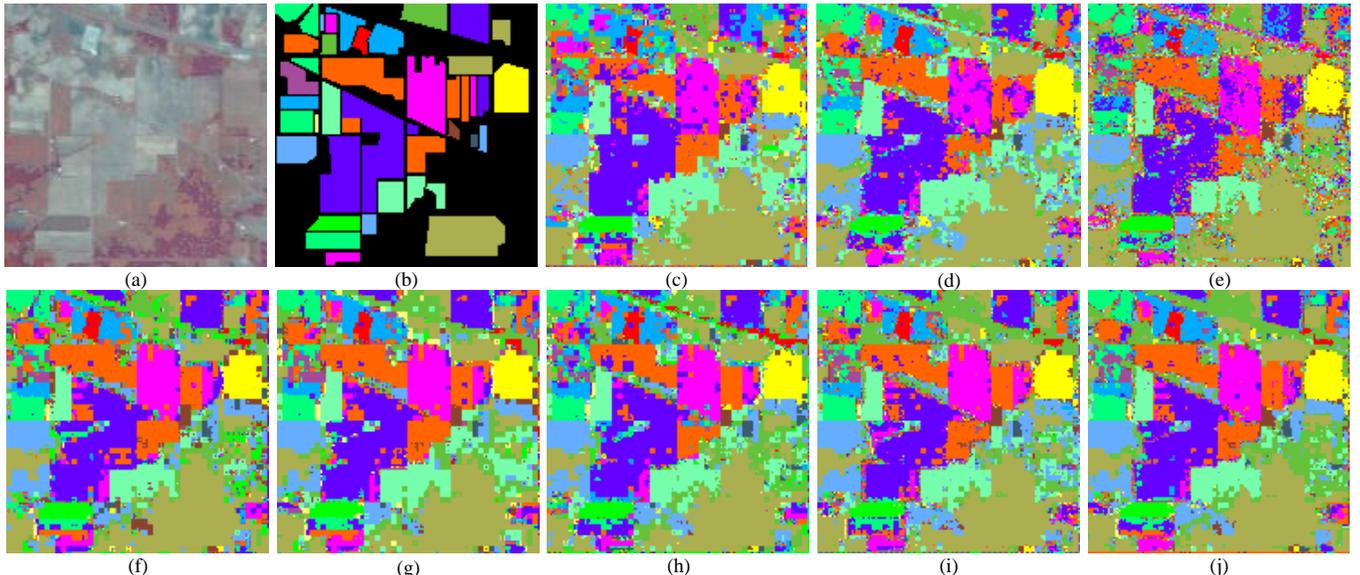

**Fig. 9.** Classification maps of different methods on the IP dataset. (a) False colour image. (b) Ground truth image. (c) SSRN. (d) DBDA. (e) SS3FCN. (f) FreeNet. (g) SSDGLt. (h) EfficientNetV2. (i) ConvNeXt. (j) High-efficiency FCN.

Before these experiments, we normalised these datasets to [−1, 1] to unify the data magnitude, thus promoting network convergence.

*B. Experimental Settings*

We compared the performance of the proposed network with that of state-of-the-art deep learning architectures, including SSRN [37], DBDA [44], spectral–spatial 3D fully convolutional network (SS3FCN) [33], FreeNet [29], SSDGL [46], ConvNeXt [51], and EfficientNetV2 [19]. Both SSRN and DBDA are patch-based 3D CNN networks. SSRN uses consecutive spectral and spatial residual blocks to learn spectral and spatial representations, respectively, followed by an average pooling layer and FC layer. DBDA includes a dense spectral branch with the channel attention module and a dense spatial branch with the position attention module. The outputs of both branches are concatenated and fed to an average pooling layer, followed by an FC layer for classification. SS3FCN considers small patches of the original HSI as input and performs pixel-to-pixel classification, where parallel 3D and 1D FCNs are used to learn joint spectral–spatial features and spectral features, respectively. FreeNet and SSDGL are both encoder–decoder-based FCN architectures. Both of them are designed to exploit global discriminative information; FreeNet uses a spectral attention-based encoder, while SSDGL uses global convolutional LSTM and a joint attention mechanism. EfficientNetV2 and ConvNeXt are state-of-the-art backbones in computer vision. EfficientNetV2 has a higher parameter efficiency and faster training speed, while ConvNeXt has comparable performance to that of transformers.

There are multiple parameters related to DL architectures. In the high-efficiency FCN, the convolution stride and size of

TABLE II
COMPARISON OF CLASSIFICATION ACCURACY OF DIFFERENT METHODS ON PU DATASET

| Metrics | SSRN | DBDA | SS3FCN | FreeNet | SSDGL | EfficientNetV2 | ConvNeXt | Proposed |
|---|---|---|---|---|---|---|---|---|
| OA (%) | 86.93 ± 1.20 | 87.35 ± 1.02 | 83.96 ± 0.69 | 85.26 ± 0.62 | 85.10 ± 0.54 | 90.41 ± 1.25 | 90.08 ± 1.05 | **91.15 ± 0.85** |
| AA (%) | 80.68 ± 2.53 | 82.05 ± 2.67 | 78.57 ± 1.56 | 81.36 ± 1.04 | 82.06 ± 1.37 | 86.67 ± 2.26 | 86.78 ± 0.83 | **89.55 ± 0.72** |
| Kappa×100 | 82.18 ± 1.54 | 82.95 ± 1.48 | 79.01 ± 0.86 | 80.22 ± 0.68 | 80.05 ± 0.73 | 87.25 ± 1.59 | 86.61 ± 1.40 | **88.16 ± 0.86** |
| C1 | 95.35 | 96.21 | 84.54 | 92.74 | 92.42 | 96.51 | 96.02 | **98.25** |
| C2 | 94.94 | 93.98 | 89.69 | 91.17 | 90.46 | 92.76 | **94.19** | 93.80 |
| C3 | 41.95 | 44.01 | 20.91 | 45.50 | 42.84 | 60.31 | 59.52 | **64.91** |
| C4 | 82.79 | 83.73 | 80.04 | **91.28** | 91.90 | 87.31 | 86.43 | 85.04 |
| C5 | 98.9 | 99.55 | 98.33 | 96.94 | 96.36 | 99.21 | 97.94 | **99.82** |
| C 6 | 68.14 | 69.83 | 86.78 | 62.14 | 60.94 | **90.20** | 70.22 | 82.08 |
| C7 | 70.38 | 72.63 | 59.16 | 70.31 | 74.30 | 63.27 | 92.48 | **93.74** |
| C8 | 90.91 | 82.04 | **93.63** | 89.47 | 92.31 | 91.49 | 88.65 | 91.58 |
| C9 | 82.86 | 96.69 | 94.13 | 92.69 | 97.01 | **98.79** | 95.58 | 96.72 |

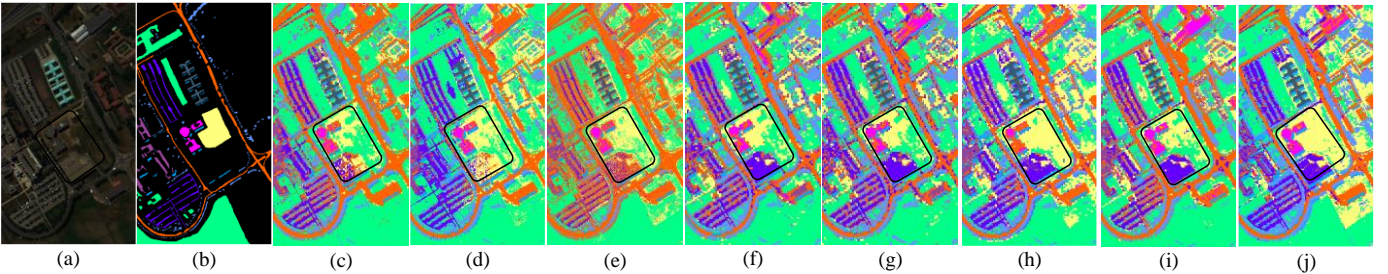

(a) (b) (c) (d) (e) (f) (g) (h) (i) (j)

**Fig. 10.** Classification maps of different methods on the PU dataset. (a) False colour image. (b) Ground truth image. (c) SSRN. (d) DBDA. (e) SS3FCN. (f) FreeNet. (g) SSDGL. (h) EfficientNetV2. (i) ConvNeXt. (j) High-efficiency FCN.

spatial padding were set to 1, while the dropout rate was set to 0.2. Other hyperparameters are presented in Fig. 4. These hyperparameters can be adjusted as per different situations. For example, the number of output channels can be halved for the PU dataset with fewer channels. For fair comparison, the above hyperparameter setting was used for all four datasets in the following experiments. The proposed network adopted the AdamW optimiser [53], where the learning rate, weight decay, and number of training epochs were set to $1 \times 10^{-4}$, $1 \times 10^{-2}$, and 150, respectively. The hyperparameters of the comparison methods were set according to the recommended values and then selected after fine-tuning to achieve the best performance. All of the methods were run on the PyTorch platform except for SS3FCN, which was implemented in Keras. All methods were trained and tested on the same sample sets generated by the proposed sampling strategy. The batch size was set to 64 for all methods. Furthermore, all experiments were conducted on a workstation with an AMD Ryzen 7 5800X 8-core processor with a 3.40 GHz CPU and an NVIDIA GeForce RTX 3060 GPU.

The classification performance was evaluated by the producer accuracy (PA) of each class, overall accuracy (OA), average accuracy (AA), and kappa coefficient (Kappa), which are formulated as follows:

$$PA_j = p_{jj} / p_{+j} \ (j=1,2,3,\cdots,K),\quad (2)$$

$$OA = \sum_{i=1}^{K} p_{ii} / N,\quad (3)$$

$$AA = \sum_{i=1}^{K} PA_i / K,\quad (4)$$

$$\text{Kappa} = \frac{N \sum_{i=1}^{K} p_{ii} - \sum_{i=1}^{K} \left( p_{i+} \times p_{+j} \right)}{N^2 - \sum_{i=1}^{K} \left( p_{i+} \times p_{+j} \right)},\quad (5)$$

where $p_{i+} = \sum_{j=1}^{K} p_{ij}$ and $p_{+j} = \sum_{i=1}^{K} p_{ij}$, $p_{ij}$ are the number of samples misclassified from the $i$-th class to the $j$-th class, and $N$ and $K$ are the number of testing samples and categories, respectively. All experiments were repeated 10 times to avoid biased estimation, and the mean values were calculated for comparison, as presented in Section IV-D.

### C. Experimental Results and Discussion

*1) Quantitative Evaluation:* Tables I–IV summarise the classification accuracy of all compared methods. We can then observe from these tables that the performance of all methods was considerably lower on the IP dataset than on the other datasets, particularly the 4-th, 9-th and 15-th categories in the IP dataset. This may be because of the lack of training data in the IP dataset and the spatial resolution of this dataset is low. Nevertheless, on all four datasets, our network achieved the highest OA, AA, and Kappa and exhibited the best or near-best accuracy in most classes. For example, on the IP dataset, the proposed method obtained the highest OA of 84.72%, which exceeded that of SSRN, DBDA, SS3FCN, FreeNet, SSDGL, ConvNeXt, and EfficientNetV2 by ~7.04%, 6.35%, 8.05%, 2.96%, 2.65%, 5.54%, and 3.64%, respectively.

Although some of the comparison methods achieved satisfactory results in previous studies, they failed to perform well on certain datasets using the proposed sampling strategy. Among these methods, SS3FCN generally exhibited the worst performance because it uses a 3D FCN and 1D FCN to learn spectral–spatial features and spectral features, respectively, thus resulting in high spectral redundancy and increased model complexity. In terms of FCN-based methods, FreeNet and SSDGL performed better on the IP dataset but worse on the other datasets. A possible reason for this is that the scarcity of labelled data made it difficult for these methods to optimally operate because these networks are more complex than others. Compared with FreeNet and SSDGL, the patch-based methods (i.e. SSRN and DBDA) performed worse on the IP dataset but better on the other three datasets. However, ConvNeXt and EfficientNetV2 exhibited good performance on all four datasets, indicating that they have superior generalisation performance. Note that the proposed network exhibited significant improvement over all the aforementioned state-of-the-art methods on all four datasets, demonstrating its effectiveness and generalisability. Even for certain indistinguishable classes (e.g. *Gravel* in the PU dataset, *railways* in the SA dataset), the proposed network classified the corresponding testing data with relatively high accuracy. These results confirm the robustness of the designed network under challenging conditions.

TABLE III
COMPARISON OF CLASSIFICATION ACCURACY OF DIFFERENT METHODS ON THE SA DATASET

| Metrics | SSRN | DBDA | SS3FCN | FreeNet | SSDGL | EfficientNetV2 | ConvNeXt | Proposed |
|---|---|---|---|---|---|---|---|---|
| OA (%) | 89.19 ± 1.86 | 88.38 ± 1.22 | 83.08 ± 1.01 | 85.43 ± 0.69 | 87.99 ± 0.96 | 89.11 ± 0.84 | 87.61 ± 1.06 | **91.90** ± 0.58 |
| AA (%) | 92.94 ± 1.62 | 90.47 ± 1.30 | 89.43 ± 0.36 | 88.96 ± 1.00 | 90.76 ± 0.89 | 92.11 ± 0.67 | 91.29 ± 0.87 | **94.78** ± 0.54 |
| $\kappa \times 100$ | 87.96 ± 2.08 | 87.06 ± 1.26 | 81.12 ± 0.96 | 83.75 ± 0.59 | 86.64 ± 1.04 | 87.83 ± 0.94 | 86.18 ± 1.18 | **90.97** ± 0.69 |
| C1 | 90.62 | 98.87 | 96.61 | **100** | 99.41 | 97.08 | 99.76 | 96.03 |
| C2 | 98.13 | 95.54 | 98.22 | 92.83 | 96.39 | 93.55 | 92.13 | **99.80** |
| C3 | 95.59 | **99.47** | 84.29 | 79.08 | 81.15 | 87.70 | 85.95 | 93.86 |
| C4 | 96.67 | 96.10 | 93.55 | 97.57 | 96.26 | 98.13 | 97.08 | **98.54** |
| C5 | 81.76 | 67.58 | **90.02** | 72.26 | 76.21 | 66.92 | 73.90 | 85.43 |
| C6 | 98.88 | 99.92 | 93.63 | 97.56 | 99.57 | 99.90 | 99.97 | **99.98** |
| C7 | **99.57** | 99.64 | 98.04 | 86.20 | 94.58 | 98.39 | 91.84 | 96.27 |
| C8 | 79.76 | 85.41 | 56.51 | 75.71 | 78.84 | **93.96** | 90.48 | 87.57 |
| C9 | 98.04 | 98.85 | 98.20 | 96.38 | 98.65 | 97.38 | 98.41 | **99.44** |
| C10 | 89.91 | 90.80 | 89.12 | 87.58 | 88.93 | 90.03 | 90.88 | **93.42** |
| C11 | 99.49 | 83.78 | 96.31 | 98.86 | 98.83 | **99.68** | 89.09 | 98.99 |
| C12 | 98.56 | 97.53 | 89.89 | 93.12 | 94.92 | 99.00 | 96.85 | **99.95** |
| C13 | 96.45 | 87.56 | 90.07 | 96.64 | 93.98 | 99.10 | 98.82 | **99.15** |
| C14 | 93.08 | 90.07 | 93.19 | 87.61 | 83.95 | **96.72** | 95.01 | 95.35 |
| C15 | 72.74 | 58.89 | 71.67 | 72.01 | 73.14 | 57.42 | 63.09 | **74.37** |
| C16 | 97.82 | 97.70 | 91.65 | 98.43 | 97.38 | **98.85** | 97.45 | 98.38 |

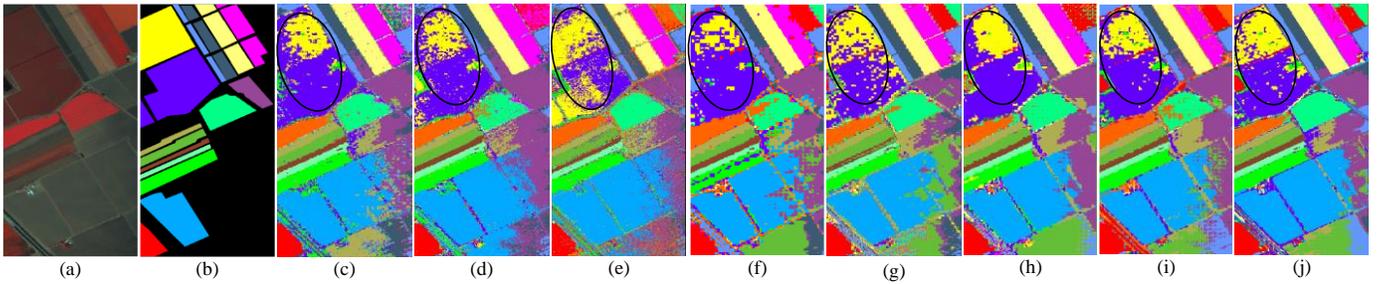

**Fig. 11.** Classification maps of different methods on the SA dataset. (a) False colour image. (b) Ground truth image. (c) SSRN. (d) DBDA. (e) SS3FCN. (f) FreeNet. (g) SSDGL-Net. (h) EfficientNetV2. (i) ConvNeXt. (j) High-efficiency FCN.

*2) Qualitative Evaluation:* Figs. 9–12 visualise the corresponding classification maps alongside the false colour images and ground truth maps. The classification maps are consistent with the reported quantitative results. For example, the classification maps produced by SS3FCN contained more noise and speckles than the maps produced by other methods on the IP, PU, and SA datasets, which agrees with the quantitative results in Tables I–IV. Among these methods, the proposed network produced the least noise and the most accurate classification maps on all datasets. Furthermore, objects covered by shadows could be identified using the proposed framework. For example, as illustrated in the black rectangles in Fig. 12, parts of buildings, roads, and vegetation were covered in shadows. SS3FCN, EfficientNetV2, ConvNeXt, and the proposed network in particular could detect shadow regions more effectively than SSRN, DBDA, FreeNet, and SSDGL. This shows the effectiveness of the designed network.

Furthermore, using the proposed sampling strategy, the class boundaries of classification maps produced by the spectral–spatial methods were more consistent with those of FCIs, particularly for the IP dataset. However, there were many scatter speckles in the classification maps because the input window size was extremely small to provide sufficient spatial information, thus resulting in inconsistent segmentations across the boundaries of windows. Therefore, selecting a larger window size is preferable when satisfying the basic principles of effective sampling strategy design, as described in Section III-A. Furthermore, the overlay inference strategy [54] can be employed to alleviate this boundary effect.

To summarise, the experiment results demonstrate the

TABLE IV
COMPARISON OF CLASSIFICATION ACCURACY OF DIFFERENT METHODS ON THE UH DATASET

| Class | SSRN | DBDA | SS3FCN | FreeNet | SSDGL | EfficientNetV2 | ConvNeXt | Proposed |
|---|---|---|---|---|---|---|---|---|
| OA (%) | 86.65 ± 1.26 | 86.96 ± 1.36 | 85.10 ± 2.01 | 84.17 ± 2.37 | 85.74 ± 1.21 | 90.51 ± 1.14 | 88.01 ± 0.69 | **91.43** ± 0.56 |
| AA (%) | 87.56 ± 1.46 | 88.50 ± 0.98 | 87.23 ± 1.56 | 82.54 ± 2.41 | 83.68 ± 0.81 | 91.77 ± 0.76 | 88.45 ± 0.69 | **92.56** ± 0.59 |
| Kappa×100 | 85.63 ± 2.06 | 85.89 ± 1.86 | 84.63 ± 2.14 | 82.40 ± 2.58 | 83.89 ± 1.01 | 89.73 ± 1.23 | 87.02 ± 0.74 | **90.73** ± 0.62 |
| C1 | 91.00 | 99.23 | 91.84 | 94.14 | 96.55 | 97.22 | 98.54 | **99.49** |
| C2 | 92.56 | 96.14 | **95.07** | 86.1 | 85.75 | 94.23 | 89.51 | 93.86 |
| C3 | 100 | **100** | 99.01 | 85.58 | 89.98 | **100** | **100** | **100** |
| C4 | 93.7 | 90.01 | 99.15 | 96.22 | 94.65 | 94.88 | 91.49 | 91.83 |
| C5 | 99.43 | **99.27** | 99.66 | 87.68 | **91.9** | 99.42 | 99.71 | 98.42 |
| C 6 | 93.19 | 97.27 | 97.76 | 86.39 | 80.91 | 98.28 | 92.98 | **98.50** |
| C7 | 87.98 | 90.83 | 74.27 | 88.23 | 84.52 | 89.60 | 85.52 | **92.73** |
| C8 | 63.28 | 62.42 | 71.92 | 59.58 | 71.96 | 69.69 | 73.32 | **76.07** |
| C9 | 83.31 | **92.26** | 88.12 | 89.22 | 91.84 | 91.87 | 91.24 | 89.16 |
| C10 | 86.40 | 84.39 | 73.08 | 69.04 | 63.10 | 91.36 | 84.99 | **93.99** |
| C11 | 70.46 | 84.36 | 73.93 | 76.07 | 69.71 | **88.26** | 75.29 | 85.72 |
| C12 | 77.70 | 58.67 | 58.40 | 67.93 | 73.87 | 77.23 | **85.07** | 82.34 |
| C13 | 81.58 | 87.09 | 92.48 | 85.40 | 81.05 | 92.39 | 87.13 | **92.55** |
| C14 | 94.64 | 86.3 | **94.94** | 73.17 | 86.47 | 92.51 | 73.41 | 94.44 |
| C15 | 98.87 | 99.34 | 98.67 | 93.31 | 93.02 | **99.71** | 98.62 | 99.35 |

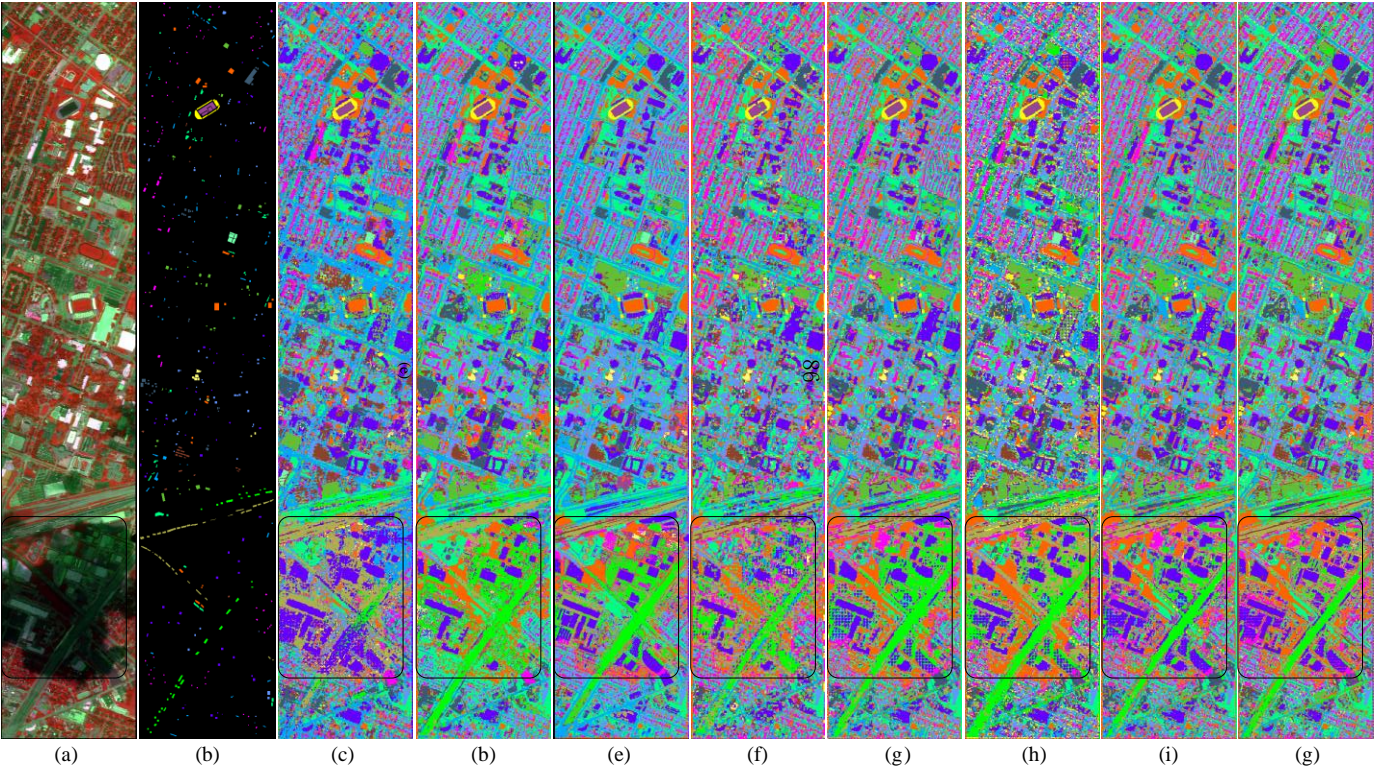

(a)  (b)  (c)  (b)  (e)  (f)  (g)  (h)  (i)  (g)

**Fig. 12**. Classification maps of different methods on the UH dataset. (a) False colour image. (b) Ground truth image. (c) SSRN. (d) DBDA. (e) SS3FCN. (f) FreeNet. (g) SSDGL. (h) EfficientNetV2. (i) ConvNeXt. (j) High-efficiency FCN.

superiority of the proposed network and indicate that the performance of spectral–spatial methods can be more accurately reflected and evaluated using the proposed sampling strategy.

## V. SENSITIVITY ANALYSIS

### A. Leakage-free Balanced Sampling Strategy Analysis

It can be seen in Figs. 5–8 that there is non-overlapping between the training and testing data and all classes exist in both training and testing data, thus demonstrating the proposed sampling strategy can avoid information leakage and achieve balanced sampling.

Moreover, we observed a trade-off between the window size and the number of windows because overly small windows resulted in limited spatial information for spatial-based methods to learn. Conversely, excessively large windows caused certain classes with limited samples to only on the training or testing set. Thus, we analysed the effect of window size on the performance of the proposed high-efficiency FCN. Because of the limited number of labelled samples for specific classes in the IP dataset (e.g. the *Oats* category had only 20 labelled pixels), we set its window size to the minimum value of 4. For other datasets, we conducted experiments to select the optimal window size. During these experiments, we varied the window size while fixing all other parameters.

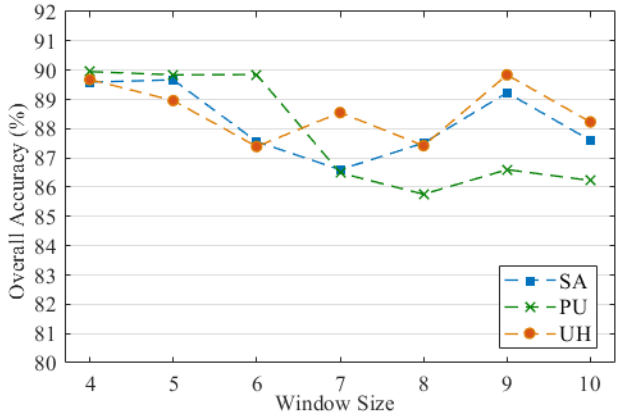

**Fig. 13.** Variation of test accuracy with input window size on the IP dataset.

TABLE V
ABLATION ANALYSIS OF THE PROPOSED HIGH-EFFICIENCY FCN ON THE IP AND UH DATASETS

| Changes | | IP | UH |
|---|---|---|---|
| (a) Baseline | - | 79.18 | 89.14 |
| (b) Activation Function | SiLU->GELU | 79.49 | 89.23 |
| | SiLU->SELU | 81.39 | 89.56 |
| (c) Normalisation | BN->LN | 81.69 | 89.69 |
| (d) Fewer LN and SELU | - | 82.47 | 89.81 |
| (e) Attention module | SE->CAM | 82.56 | 89.79 |
| | Remove SE | 82.72 | 89.84 |
| (f) Inverted channel | - | 84.72 | 91.43 |

Unlike patch-based classification, where the accuracy improves as the patch size increases, in our experiments, the accuracy did not increase with increase in the window size, as illustrated in Fig. 13. In fact, there was almost no direct relationship between the accuracy and window size. Moreover, the difference in accuracy for different window sizes was minor, again demonstrating that the proposed sampling strategy can eliminate the spatial dependence between the training and testing data.

Although the smallest window size achieved the highest accuracy on certain datasets, it failed to provide sufficient spatial information for methods with strong spatial information extraction ability. Moreover, smaller window sizes resulted in lower inference efficiency and more scatter points (Fig. 9). Therefore, selecting a larger window size with a comparable accuracy was preferable. Weighting the efficiency and accuracy, we set the window size to 6 for the PU dataset, 9 for the SA dataset, and 9 for the UH dataset.

*B. High-Efficiency FCN Analysis*

We then analysed the proposed network design by following a trajectory from EfficientNetV2 to the high-efficiency FCN. Experiments were conducted on the IP and UH datasets, both of which are typical and challenging datasets. The corresponding results are summarised in Table V.

The normalisation layer and activation function are important components of a CNN. We first evaluated the performance of the proposed network with different activation functions, including SiLU, SELU, and GELU. SiLU and GELU are used in EfficientNetV2 [19] and ConvNeXt [51], respectively. SELU possesses self-normalising properties that allow neural network learning to be highly robust. As shown in Table V(b), the network trained with GELU achieved the best results with an OA of 81.39% on the IP dataset and an OA of 89.56% on the UH dataset. Therefore, the proposed network adopted GELU as the activation function and used it in the following experiments.

For normalisation, as illustrated in Table V(c), our network trained with LN resulted in a slightly higher OA compared with BN, with gains of 3% and 1% on the IP and UH datasets, respectively. Therefore, we used LN for normalisation in our proposed network.

The number of activation and normalisation layers has an impact on the performance of networks. As illustrated in Table V(d), after reducing the number of LN and SELU activation layers, the classification accuracy on both datasets did not decline but slightly improved. The reason for this may be that SELU induces self-normalising properties; thus, it is not necessary to perform normalisation again.

To avoid overfitting and reduce the number of trainable parameters, we used the channel attention module [18] with no trainable parameters to replace the SE module, which did not contribute to the accuracy. Then, we attempted to remove the attention module from our network. Interestingly, this led to marginal improvements on both datasets (from 82.47% to 82.72% on the IP dataset and from 89.81% to 89.84% on the UH dataset). Thus, our network did not contain an attention module.

As detailed in Table V(f), the *inverted channels* led to a significant increase in the OA from 82.72% to 84.72% on the IP dataset and from 89.84% to 91.43% on the UH dataset. This demonstrates that the *inverted channels* setting can help the network to mine additional discriminative spectral information.

The modified network improved the classification results for both datasets. The superior performance of our network is considered to be attributed to the network's better ability to mine valuable information from the redundant spectral bands.

*C. Model Complexity and Speed Analysis*

To comprehensively analyse the complexity of the proposed network, we calculated the number of trainable parameters (Params) and FLOPs as well as the training and inference time for the comparison methods on the IP and UH datasets. Params and FLOPs are indirect measures of the computational complexity, whereas the runtime is a direct measure.

As shown in Table VI, the proposed method generally achieved the best results, particularly in terms of the training and inference time, and near-best results in terms of Params and FLOPs. SS3FCN, FreeNet, and SSDGL had more Params than other methods. Although SS3FCN had the least FLOPs, its training and inference time was the longest since it not only employed 3D networks with many Params but also used a three-time prediction averaging strategy. Compared to the patch-based methods (i.e. SSRN and DBDA), the FCN-based methods (except for SS3FCN) required less time for inference. Note that the time-consuming training process is conducted offline; however, the inference speed is the main factor determining whether a method is practical. Thus, the pixel-to-pixel classification strategy is more suitable for practical applications. The proposed network had the fastest inference speed among the compared networks.

TABLE VI
COMPARISON OF PARAMS, FLOPS, TRAINING AND INFERENCE TIME OF DIFFERENT METHODS ON IP AND UH DATASETS

| Dataset | Metric | SSRN | DBDA | SS3FCN | FreeNet | SSDGL | EfficientNetV2 | ConvNeXt | Proposed |
|---|---|---|---|---|---|---|---|---|---|
| IP | Params | 407K | 447K | 2286K | 2648K | 2266K | 455K | 1034K | **373K** |
| | FLOPs | 1,133,540K | 762,899K | **4,571K** | 1,602,282K | 3,522,861K | 230,092K | 593,768K | 215,138K |
| | Training (s) | 398 | 486 | 2,555.53 | 387 | 4952 | 604 | 592 | **365** |
| | Inference (s) | 2.21 | 4.32 | 2.59 | 0.44 | 2.18 | 0.33 | 0.57 | **0.16** |
| UH | Params | **331K** | 344K | 4,320K | 2,599K | 2153 | 443K | 1,009K | 366K |
| | FLOPs | 5,791,167K | 3,891,008K | **8,638K** | 11,188,779K | 24,581,281 | 1,561,525K | 4,122,603K | 1,500,381K |
| | Training (s) | 13,640 | 14,123 | 29,641 | 8,228 | 54,323 | 8,649 | 10,325 | **6,083** |
| | Inference (s) | 5.12 | 6.32 | 5.16 | 0.69 | 4.15 | 0.53 | 0.56 | **0.37** |

### D. Impact of the Number of Training Samples

To test the robustness and stability of the proposed method, we performed experiments on fewer training samples per class on the IP dataset. This dataset is a typical unbalanced dataset with extremely few labelled samples, thus posing significant challenges to supervised methods. Fig. 14 shows the OA of different methods with different numbers of training samples, where the training percent represents the proportion of training samples in Fig. 5(a). For example, 100% is the total number of training samples, as listed in Fig. 5(a). For all methods, the accuracy decreased with fewer training samples, particularly when the training percent was <50%. Nonetheless, the proposed network consistently outperformed the other methods in terms of accuracy, thus demonstrating its robustness.

## VI. CONCLUSION

This study proposes an OEEL framework for tiny HSI datasets to facilitate efficient classification and objective performance evaluation. In this framework, the proposed leakage-free balanced sampling strategy can generate balanced training and testing samples without overlap or information leakage, thus enabling objective performance evaluation. Based on the generated samples, a high-efficiency FCN is proposed to avoid redundant computation while showing a favorable accuracy-speed trade-off. Both the quantitative and qualitative experimental results demonstrate that the proposed high-efficiency FCN outperformed many state-of-the-art methods, including SSRN, DBDA, SS3FCN, FreeNet, SSDGL, EfficientNetV2, and ConvNeXt.

However, the experimental results in this study may fail to identify suitable DL-based architectures because the lack of HSI datasets prevents some of these architectures from realising their full potential. Therefore, future work should construct large benchmark datasets to facilitate future spectral–spatial HSI analysis research. Furthermore, we will consider weakly supervised approaches to relieve the demand for expensive pixel-level image annotation.

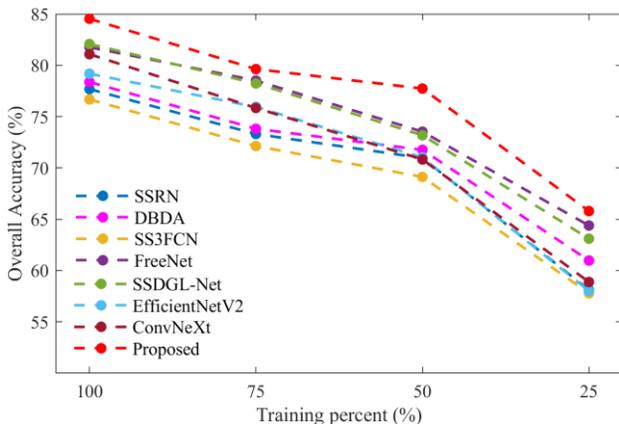

**Fig. 14.** The classification accuracy of different methods with a varying number of training samples on the IP dataset.